\pdfoutput=1
\documentclass[11pt]{article}
\usepackage[final]{acl}
\usepackage{times,latexsym}
\usepackage[T1]{fontenc}
\usepackage{microtype,courier,graphicx}
\usepackage{booktabs,array,amsmath,amssymb,url,xcolor}
\definecolor{ResultBlue}{RGB}{0,70,170}

\title{BASIL-W: Scope-Aware Evidence for Text-to-SQL Selection}
\author{%
  \textbf{Anup Roy}$^{1}$ \quad
  \textbf{Rishabh Gyanendra Upadhyay}$^{1}$ \quad
  \textbf{Animesh Rameshbhai Panara}$^{1}$ \\
  \textbf{Aidan Philip Millar}$^{2}$ \quad
  \textbf{Larry Murray}$^{1}$ \quad
  \textbf{Robin Mills}$^{1}$ \\[4pt]
  $^{1}$Inception42, Abu Dhabi, UAE \qquad
  $^{2}$Mubadala Investment Company, Abu Dhabi, UAE \\[2pt]
  \small\texttt{\{anup.roy, rishabh.upadhyay, larry.murray, robin.mills\}@inceptionai.ai},
  \small\texttt{amillar@mubadala.ae}
}
\begin{document}
\maketitle
\begin{abstract}
Execution feedback can mislead text-to-SQL selectors when incomplete query
outputs are treated as conclusive evidence. BASIL-W represents observations
as \emph{scope-aware witnesses} containing the executed SQL, its output,
completion status, and certified scope. An absence claim requires a
completed check covering the claimed predicate and domain. A $2\times2$
factorial experiment varies scope metadata and an explicit instruction while
holding the model, candidates, and probes fixed. On the full Spider and
WikiSQL test splits with GPT-4.1, execution accuracy increases from 85.98\%
to 88.50\% and from 90.50\% to 91.60\% (all six paired contrasts
Holm-adjusted $p<0.006$). The combined interface outperforms either factor
alone. The arm ordering also holds with Qwen2.5-Coder-32B and Llama-3.3-70B
and on BIRD dev, where the gain reaches $+3.7$ percentage points. Observed
margins over the strongest matched selection baseline are $+1.58$ on Spider,
$+1.76$ on BIRD dev, and $+0.15$ on WikiSQL; these comparisons lack paired
uncertainty estimates. The Spider audit found unsupported absence-claim
rates of 12.0\% for raw evidence and 4.3\% for BASIL-W. The design does not isolate scope content from prompt length or
presentation, and the audit does not establish a causal mechanism.

\end{abstract}
\section{Introduction}
\begin{figure*}[t]
\centering
\includegraphics[width=\textwidth, trim={0 0 0 2.2cm}, clip]{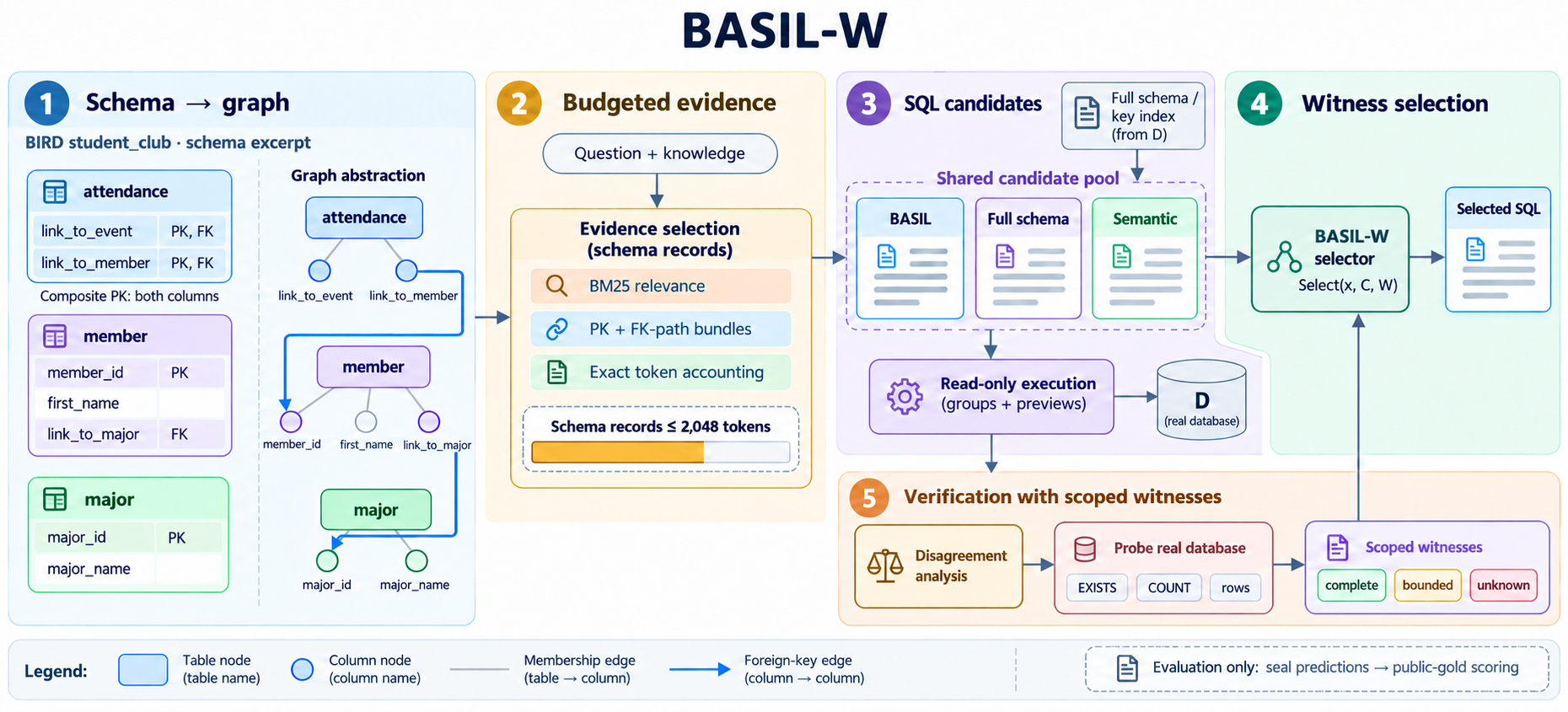}
\caption{BASIL-W generation and selection pipeline. BASIL denotes the
graph-guided candidate branch; BASIL-W adds scope-aware selection. The
2,048-token limit applies to selected schema records only. The factorial
comparison intervenes at the evidence interface while preserving candidate
queries and observations.}
\label{fig:architecture}
\end{figure*}

A text-to-SQL selector may reject a correct query for the wrong reason: a
department name missing from a twelve-row preview may still exist in the
table, and treating its omission as a database fact can favor an incorrect
candidate. A completed existence check answers a different question, and
reliable selection requires preserving that difference as execution output
enters the model context. BASIL-W makes the boundary explicit: a database
witness carries the executed SQL, its output, the execution outcome, and
what the output covers. The selector chooses from a fixed candidate pool
without modifying the candidate SQL (Figure~\ref{fig:architecture}).

Multi-prompt selection \citep{mcs}, complementary generation \citep{chase},
execution-based verification \citep{dpc}, and graph schema retrieval
\citep{schemagraph} provide related approaches. Our contribution is
\emph{qualifying the evidence used to discriminate between existing
candidates}, evaluated through a $2\times2$ factorial experiment. Across the
evaluated settings spanning three backbones and three benchmarks, the
combined interface improves execution accuracy by 1.0 to 3.7 points over raw
evidence, more than either factor alone. The Spider audit found unsupported
absence-claim rates of about 12\% for raw evidence and 4\% for BASIL-W under
the predicate-matched support criterion.

\section{Scope-Aware Evidence}
\paragraph{Task and candidate pool.}
Given a question $x$ and database $D$, the generator produces three
candidates $C=\{q_1,q_2,q_3\}$ using graph-guided generation with
execution-guided refinement, full-schema generation, and semantic
composition. Graph packing ranks schema records by BM25 under a 2,048-token
budget. These components are shared across arms. Reference labels are
withheld during generation, probing, and selection and used only for
evaluation. A shared
planner issues up to six read-only probes (five-second limits, twelve-row
previews) and, for resolvable column--literal equality predicates, up to
eighteen parameterized existence checks; all probes and outputs are cached
once for every arm. Each selector returns a candidate ID plus a structured
rationale citing evidence IDs and any explicit absence assertions; the ID is
validated (out-of-pool queries fail), and the audit assesses the explicit
assertions in the rationale.

\paragraph{Witness contract.}
Let $D$ be a fixed database snapshot. An observation is
$o=(q_o,z_o,e_o,c_o)$: the executed probe $q_o$ with bound parameters, its
execution output $z_o=\operatorname{Exec}(q_o,D)$, the outcome
$e_o\in\{\text{ok},\text{timeout},\text{error}\}$, and completion
$c_o\in\{0,1\}$ ($c_o=1$ iff the output cursor was exhausted). The output
$z_o$ is distinct from source rows: \textsc{exists} returns a Boolean
scalar, a count a number, and a scan a tuple list. A claim fixes a relation
expression $U$, including its joins and domain filters, and a SQL predicate
$\phi$. Write $Q_\phi(D)=\sigma_{\phi=\mathrm{TRUE}}U(D)$ and
$\alpha_\phi\equiv Q_\phi(D)=\varnothing$.

Coverage is a query-level certificate $K_o=(U,\phi)$, issued only when
$\operatorname{Covers}(q_o;U,\phi)$ is verified from the probe template,
SQL and bound parameters, independently of which rows exist. The
conservative rule accepts an exact \textsc{exists} or ungrouped
\texttt{COUNT(*)} check over $Q_\phi$, or a scan of the same $U$ with filter
$\psi\in\{\mathrm{TRUE},\phi\}$ and a projection retaining the columns
needed to evaluate $\phi$. No \texttt{LIMIT}, \texttt{OFFSET}, grouping or
\texttt{HAVING} may restrict the certified search; other query forms receive
no certificate. Matching preserves literals, parameters, types, collations
and SQL NULL semantics. A preview stopped at the cap has $c_o=0$; an
exhausted \texttt{LIMIT} query can have $c_o=1$ without a certificate.
Completion concerns the probe's output, so a completed \textsc{exists}
check need not visit every source row.

For certified templates, $\operatorname{Neg}(o,\phi)$ means a scalar zero
for \textsc{exists}/\texttt{COUNT(*)}, or no returned scan tuple on which
$\phi$ evaluates to SQL TRUE. Table~\ref{tab:rules} gives the licensed
interpretations; $\alpha_\phi$ is \emph{supported} by observations $O$ iff
\begin{equation}
\begin{split}
 \exists o\in O:\quad &e_o=\text{ok}\ \land\ c_o=1\\
 &\land\ \operatorname{Covers}(q_o;U,\phi)\\
 &\land\ \operatorname{Neg}(o,\phi).
\end{split}
\label{eq:support}
\end{equation}
\begin{table}[t]
\centering\fontsize{8.2}{9.6}\selectfont
\setlength{\tabcolsep}{3pt}\renewcommand{\arraystretch}{1.05}
\begin{tabular}{@{}p{0.44\columnwidth}p{0.52\columnwidth}@{}}
\toprule
Observation and execution status & Licensed interpretation \\
\midrule
\textsc{exists} over exactly $Q_\phi$ & $1$: presence; $0$: $Q_\phi(D)=\varnothing$ \\
Ungrouped \texttt{COUNT(*)} over $Q_\phi$ & $>0$: presence; $0$: absence \\
Scan with a verified matching tuple & presence, within that tuple's provenance \\
Scan without a matching tuple & absence only if $\operatorname{Covers}(q_o;U,\phi)$ and $c_o=1$; else unknown \\
Error, timeout, or unverified match between probe and claim & unknown \\
\bottomrule
\end{tabular}
\caption{Absence requires successful completion and a verified match to
the claimed predicate and domain. A matching tuple can
establish presence even in an incomplete scan.}
\label{tab:rules}
\end{table}

Thus a completed check of \texttt{name = 'Sales'} cannot support absence
of \texttt{name = 'Finance'}, even when Finance is absent. An empty join
result does not imply that either input is empty, and outer-join padding
does not establish a match in the padded input. \texttt{COUNT(col)}$=0$
does not imply that no rows exist, because NULL values are excluded;
\texttt{col = NULL} does not test \texttt{col IS NULL}. The contract is
template-bound; we claim no general SQL scope inference. The gap between
answer completeness and source completeness is classical in database
theory \citep{motro1989,levy1996,razniewski2011}; we apply this distinction
to the evidence presented to an LLM selector. The raw arm serializes $(q_o,z_o,e_o)$;
the witness adds $c_o$, certified scope $K_o$ (or no certificate), and the
licensed interpretation. The contract specifies the claims an observation
supports; adherence by the model is evaluated empirically.

\paragraph{Predicate-matched evidence on Spider.}
The published \texttt{concert\_singer} development question asks for the
average, minimum and maximum ages of singers from France; its reference SQL
uses \texttt{country = 'france'} \citep{spiderexamples}. For an absence
claim about that predicate on \texttt{singer}, the matching probe is
\begin{quote}
\small\ttfamily\raggedright
SELECT EXISTS (SELECT 1 FROM singer\\
WHERE country = 'france');
\end{quote}
Successful completion with output $0$ supports the claim; output $1$
establishes presence. A completed check of \texttt{country = 'japan'}
cannot support it, and neither can \texttt{SELECT country FROM singer
LIMIT 12}, even if no displayed value matches. An exhausted, unfiltered
\texttt{SELECT country FROM singer} can support absence if no returned
value satisfies the predicate. These conditions concern the development
question; the 400-question test audit is described separately below.

\paragraph{Illustrative failure case.}
Figure~\ref{fig:witness} shows the same two observations under both
interfaces: a capped scan ($c_o=0$) omits a value stored beyond the cap,
while the completed check, which both arms receive, finds it.
Execution-majority voting groups candidates by their returned results.
DPC \citep{dpc} instead constructs a distinguishing database and checks SQL
against an independent Python/Pandas solution. BASIL-W addresses the
interpretation of cached observations by making their scope and completion
explicit.
\begin{figure}[t]
\centering
\fontsize{7.2}{8.4}\selectfont\ttfamily
\begin{tabular}{@{}p{0.47\columnwidth}@{\hspace{5pt}}p{0.49\columnwidth}@{}}
\toprule
\rmfamily\bfseries Raw observation ($A_{00}$) & \rmfamily\bfseries Scope-aware witness ($A_{11}$) \\
\midrule
probe p1: SELECT name FROM dept ORDER BY id \newline
rows shown (preview cap 12): Sales, HR, Legal, \ldots, Audit &
query: p1 (same SQL, no LIMIT) \newline
outcome: ok \newline
boundary: stopped at preview cap 12; cursor not exhausted \newline
result\_complete: no \newline
covers(dept): no \newline
absence\_established: no \\
\addlinespace[3pt]
probe p2: SELECT EXISTS(SELECT 1 FROM dept WHERE name = ?) [`Finance'] \newline
rows (1): 1 &
check: p2 (same SQL and parameter) \newline
outcome: ok \newline
result\_complete: yes \newline
scope: predicate dept.name = `Finance' \newline
result: true (present) \\
\midrule
\multicolumn{2}{@{}p{0.98\columnwidth}@{}}{\rmfamily Instruction ($I{=}1$): \emph{Absence requires a completed check
covering the claimed predicate; a preview or timeout is insufficient.}} \\
\bottomrule
\end{tabular}
\caption{Illustrative serialization of the same two cached observations for the raw arm (left)
and BASIL-W (right): identical SQL, parameters and output; only annotations
and the instruction differ. The planner stops reading p1 at the twelve-row
cap, so the result is incomplete and does not cover \texttt{dept}, where
Finance sits beyond row twelve (with \texttt{LIMIT 12} it would be complete
yet still not cover the table). The completed check p2 finds Finance.
Field names are abbreviated for readability.}
\label{fig:witness}
\end{figure}

\paragraph{Factorial intervention.}
Let $M\in\{0,1\}$ enable structured scope metadata and $I\in\{0,1\}$ enable a
single explicit instruction: absence requires a completed check covering the
claimed predicate; a preview or timeout is insufficient. The four arms are
$A_{00}$ (raw), $A_{10}$ (metadata), $A_{01}$ (instruction), and $A_{11}$
(BASIL-W). Each arm is a selector $s_{MI}(x,C,\tau_{MI}(O))\in\{1,2,3\}$ over the same
question $x$, pool $C$ and observation set $O$; only the serialization
$\tau_{MI}$ differs, with $\tau_{00}$ rendering $(q_o,z_o,e_o)$ and
$\tau_{1I}$ adding $c_o$, $K_o$ and the licensed interpretation, while
$I$ toggles the instruction. Metadata is
derived from the stored trace without new queries, so $O$ is byte-identical
across arms. Contrasts evaluate the controlled change in $\tau$, which
includes scope annotations, prompt length and presentation; the design
does not identify their separate contributions. $A_{11}-A_{00}$ is the joint contrast,
$A_{11}-A_{01}$ and $A_{11}-A_{10}$ estimate each factor's effect with the
other present, and
$\gamma=(A_{11}-A_{01})-(A_{10}-A_{00})$ quantifies departure from additivity.

\begin{table*}[t]
\centering\fontsize{8.6}{10}\selectfont
\setlength{\tabcolsep}{3.6pt}\renewcommand{\arraystretch}{1.04}
\begin{tabular}{@{}llrlrrrr@{}}
\toprule
Test & Arm & Correct & EX (\%) [95\% CI] & $\Delta$ raw & Unsupp.\,/\,claims & W\,/\,L & Holm $p$ \\
\midrule
Spider  & Raw ($A_{00}$)         & 1,846 & 85.98 [84.45, 87.38] & ref.      & 72/600 (12.00\%) & 90 / 36 & $4.9\times10^{-6}$ \\
Spider  & Metadata ($A_{10}$)    & 1,870 & 87.10 [85.61, 88.45] & $+1.12$   & 43/560 (7.68\%)  & 70 / 40 & $5.4\times10^{-3}$ \\
Spider  & Instruction ($A_{01}$) & 1,862 & 86.73 [85.22, 88.10] & $+0.75$   & 52/580 (8.97\%)  & 78 / 40 & $1.2\times10^{-3}$ \\
Spider  & \textbf{BASIL-W} ($A_{11}$) & \textbf{1,900} & \textbf{88.50} [87.08, 89.78] & $\mathbf{+2.52}$ & \textbf{22/510 (4.31\%)} & --- & --- \\
\midrule
WikiSQL & Raw ($A_{00}$)         & 14,370 & 90.50 [90.04, 90.95] & ref.    & 220/2,000 (11.00\%) & 260 / 86 & $1.1\times10^{-20}$ \\
WikiSQL & Metadata ($A_{10}$)    & 14,433 & 90.90 [90.44, 91.34] & $+0.40$ & 155/1,910 (8.12\%)  & 190 / 79 & $4.1\times10^{-11}$ \\
WikiSQL & Instruction ($A_{01}$) & 14,417 & 90.80 [90.34, 91.24] & $+0.30$ & 177/1,950 (9.08\%)  & 210 / 83 & $3.7\times10^{-13}$ \\
WikiSQL & \textbf{BASIL-W} ($A_{11}$) & \textbf{14,544} & \textbf{91.60} [91.16, 92.02] & $\mathbf{+1.10}$ & \textbf{79/1,800 (4.39\%)} & --- & --- \\
\bottomrule
\end{tabular}
\caption{Factorial comparison on the full Spider ($n=2{,}147$) and WikiSQL
($n=15{,}878$) test splits with GPT-4.1; all arms share model, candidate pool
and cached observations. EX has Wilson 95\% intervals; $\Delta$ is the gain
over raw evidence in percentage points. W/L counts BASIL-W wins/losses
against the arm in that row; Holm $p$ tests that same paired contrast,
adjusted across all six comparisons. Unsupp.\,/\,claims is unsupported
over audited absence assertions (Spider: seeded 400-question sample;
WikiSQL: descriptive counts over collected assertions from an unrecorded
sample, not population estimates).}
\label{tab:factorial}
\end{table*}

\section{Evaluation Design}
\paragraph{Benchmarks and access.}
The primary benchmarks are the released Spider 1.0 test split (2,147
questions) \citep{spider1,spiderpublic} and WikiSQL's 15,878-question test
split, a single-table diagnostic \citep{wikisql,sqlova,wikisqlpublic}.
BIRD dev \citep{bird} (1,534 questions, larger databases) is a secondary
evaluation and is identified as a development split throughout.
Database contents are available to every arm, which differs from WikiSQL's
schema-only leaderboard.

\paragraph{Models, controls, and scoring.}
The primary backbone was GPT-4.1 through an Azure-compatible endpoint (API
version \texttt{2024-10-21}); the four arms were also evaluated with
Qwen2.5-Coder-32B-Instruct and Llama-3.3-70B-Instruct. Within each setting,
the model, decoding settings, candidate pool, preview length, and output cap
were held fixed across arms. Task and cluster IDs were fixed before
scoring, and predictions were sealed before reference answers were read.
The shared candidate pools and observations were cached once and reused
across arms. Complete run manifests and the prompt-tuning history are not
included in the accompanying package (Limitations).
Spider and BIRD use original-database EX \citep{testsuite}; WikiSQL uses
its structured-query evaluator. Failures and timeouts remain in the
denominator.

\paragraph{Selection headroom.}
Let $o_i=\max_{q\in C_i}\mathbf1[q\text{ correct}]$ and let $y_i^a$ denote
selector correctness. A fixed-pool selector satisfies $y_i^a\leq o_i$, and
the errors decompose exactly as $1-y_i^a=(1-o_i)+(o_i-y_i^a)$: generation
misses plus selection misses. Oracle@3 minus raw EX is the selection headroom.

\paragraph{Paired inference.}
For $n$ tasks, paired wins $b$ and losses $c$ give $\Delta=100(b-c)/n$.
The primary family comprises the three BASIL-W contrasts on each GPT-4.1
test split. Exact two-sided McNemar tests use Holm adjustment across these
six contrasts; the tests treat questions as independent and do not account
for database clustering (Limitations). Marginal EX uses Wilson intervals.
The interaction
$\gamma=\mathrm{EX}_{11}-\mathrm{EX}_{10}-\mathrm{EX}_{01}+\mathrm{EX}_{00}$
is reported as a point estimate only. Paired uncertainty estimates are not
reported for the selection baselines or the open-weight and BIRD settings.

\paragraph{Selection baselines.}
Four additional selectors use the same candidate pools with GPT-4.1
(Table~\ref{tab:baselines}). The USC-style selector used
Universal Self-Consistency \citep{usc} to choose the candidate judged most
consistent with the others, without execution observations. The matched
judge performed the same candidate-choice task with probe observations
withheld. Execution-majority voting selected from the largest group of
candidates with matching execution results; the MCS-SQL-style selector used
a multiple-choice prompt adapted to the fixed three-candidate pool
\citep{mcs}. Evidence-based selectors received the same cached observations
as the factorial arms. The comparison evaluates these selection-stage
adaptations rather than the complete generation pipelines of the named
systems.

\paragraph{Evidence-error audit.}
We verified the Spider and WikiSQL audit statistics at the assertion level
under the predicate-matched criterion in Equation~\eqref{eq:support}.
The Spider audit covered all explicit absence assertions from the four
arms on 400 questions sampled without replacement before output inspection.
Two annotators, blinded to arm identity, assessed each assertion using its
text, claimed predicate and domain, cited evidence IDs, SQL and bound
parameters, outputs, execution outcome, and completion status. An assertion
was supported when its cited evidence satisfied the rule, unsupported when
it failed, and unresolved when the predicate or required evidence could
not be recovered. Annotations were assigned independently before
adjudication; agreement was measured on the independent labels.
Figure~\ref{fig:results}b and Tables~\ref{tab:factorial} and~\ref{tab:audit}
report the verified statistics. The WikiSQL audit describes the collected
assertions only: its original sampling procedure and number of distinct
questions were not recorded, so it does not support population estimates.

\section{Results}
\label{sec:results}
\begin{table}[t]
\centering\fontsize{8.4}{9.8}\selectfont
\setlength{\tabcolsep}{2.8pt}\renewcommand{\arraystretch}{1.05}
\begin{tabular}{@{}lrrr@{}}
\toprule
Selector (same pool, GPT-4.1) & Spider & WikiSQL & BIRD dev \\
\midrule
USC-style, no observations      & 84.54 & 89.94 & 63.89 \\
Matched judge, no probes        & 85.42 & 90.44 & 65.06 \\
Raw evidence ($A_{00}$)         & 85.98 & 90.50 & 65.71 \\
Execution-majority voting       & 86.35 & 91.45 & 66.30 \\
MCS-SQL-style selection prompt  & 86.91 & 91.37 & 67.67 \\
\textbf{BASIL-W} ($A_{11}$)     & \textbf{88.50} & \textbf{91.60} & \textbf{69.43} \\
\midrule
$\Delta$ vs.\ execution-majority & $+2.14$ & $+0.15$ & $+3.13$ \\
$\Delta$ vs.\ strongest baseline & $+1.58$ & $+0.15$ & $+1.76$ \\
\bottomrule
\end{tabular}
\caption{Selection baselines using the same candidate pools. Execution accuracy is reported as percentages, with differences in percentage points.}
\label{tab:baselines}
\end{table}

\begin{table}[t]
\centering\fontsize{8.4}{9.8}\selectfont
\setlength{\tabcolsep}{2.1pt}\renewcommand{\arraystretch}{1.05}
\begin{tabular}{@{}lrrrrr@{}}
\toprule
Arm & Claims & Unsupp. & Unres. & $\kappa$ & Questions \\
\midrule
Raw         & 600 & 72 (12.00\%) & 18 & 0.872 & 59/400 (14.8\%) \\
Metadata    & 560 & 43 (7.68\%)  & 16 & 0.845 & 38/400 (9.5\%) \\
Instruction & 580 & 52 (8.97\%)  & 17 & 0.821 & 46/400 (11.5\%) \\
\textbf{BASIL-W} & 510 & \textbf{22 (4.31\%)} & 14 & 0.809 & \textbf{20/400 (5.0\%)} \\
\bottomrule
\end{tabular}
\caption{Spider absence-claim audit, GPT-4.1, 400 questions sampled by seed
before inspection: all absence assertions per arm, those adjudicated
unsupported or unresolved, Cohen's $\kappa$ between blinded annotators (raw
agreement 95.9--97.7\%), and questions with an unsupported assertion.}
\label{tab:audit}
\end{table}

\begin{table*}[t]
\centering\fontsize{8}{9.2}\selectfont
\setlength{\tabcolsep}{3pt}\renewcommand{\arraystretch}{1.0}
\begin{tabular}{@{}llllrr@{}}
\toprule
Test & System & Backbone / strategy & Training & EX (\%) & $\Delta$ (pp) \\
\midrule
Spider & SIRIUS-SQL \citep{sirius} & 32B specialist + Gemini 3.1 Pro & fine-tuned & 91.20 & $-2.70$ \\
Spider & DeepEye-SQL \citep{deepeye} & Qwen3-Coder-30B-A3B, SE-style workflow & prompt-only & 89.8 & $-1.30$ \\
Spider & OmniSQL + MV \citep{omnisql} & OmniSQL-32B, majority voting & fine-tuned & 89.8 & $-1.30$ \\
Spider & XiYan-SQL \citep{xiyan} & Qwen2.5-Coder ensemble + GPT-4o & fine-tuned & 89.65 & $-1.15$ \\
Spider & MCS-SQL \citep{mcs} & GPT-4, multi-prompt, multiple-choice & prompt-only & 89.6 & $-1.10$ \\
Spider & Arctic-Text2SQL-R1 \citep{arctic} & 32B, RL-trained, greedy & RL & 88.7 & $-0.20$ \\
Spider & MCI-SQL \citep{mci} & nine candidates, intermediate correction & prompt-only & 88.7 & $-0.20$ \\
Spider & \textbf{BASIL-W (ours)} & GPT-4.1, three candidates, scoped selection & prompt-only & \textbf{88.50} & --- \\
Spider & CHASE-SQL \citep{chase} & Gemini 1.5, multi-path, trained selector & trained selector & 87.6 & $+0.90$ \\
Spider & OpenSearch-SQL \citep{opensearch} & GPT-4o, dynamic few-shot, consistency & prompt-only & 87.1 & $+1.40$ \\
\midrule
WikiSQL & SeaD + EG \citep{sead} & schema-aware denoising, EG decoding & supervised & 93.0 & $-1.40$ \\
WikiSQL & HydraNet + EG \citep{hydra} & hybrid ranking network, EG & supervised & 92.2 & $-0.60$ \\
WikiSQL & \textbf{BASIL-W (ours)} & GPT-4.1, zero-shot, content access & prompt-only & \textbf{91.60} & --- \\
WikiSQL & SQLova + EG \citep{sqlova} & BERT parser, EG & supervised & 89.6 & $+2.00$ \\
\bottomrule
\end{tabular}
\caption{Published test scores next to BASIL-W: reference points, not matched
comparisons (backbones, candidate budgets, training and WikiSQL access
differ). DIN-SQL and DAIL-SQL report test-suite accuracy and are omitted.}
\label{tab:published}
\end{table*}

\begin{figure*}[t]
\centering
\includegraphics[width=0.90\textwidth]{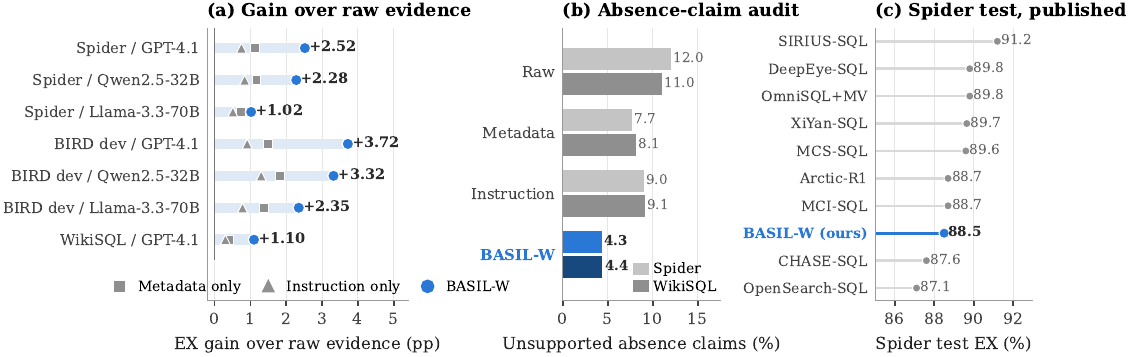}
\caption{(a) EX gain over raw evidence per arm (Table~\ref{tab:backbones}).
(b) Unsupported absence-claim rates, GPT-4.1; WikiSQL counts
describe the collected assertions only. (c) BASIL-W
beside published Spider test scores (Table~\ref{tab:published}).}
\label{fig:results}
\end{figure*}
\begin{table*}[t]
\centering\fontsize{8.3}{9.8}\selectfont
\setlength{\tabcolsep}{3pt}\renewcommand{\arraystretch}{1.04}
\begin{tabular}{@{}llrrrrrrrrr@{}}
\toprule
Benchmark & Backbone & Raw & Metadata & Instruction & BASIL-W & $\Delta$ raw & Oracle@3 & Headroom & Recovered & $\hat\gamma$ \\
\midrule
Spider test & GPT-4.1 & 85.98 & 87.10 & 86.73 & \textbf{88.50} & $+2.52$ & 93.99 & 8.01 & 31.4\% & $+0.65$ \\
Spider test & Qwen2.5-Coder-32B & 82.25 & 83.42 & 83.09 & \textbf{84.54} & $+2.28$ & 90.13 & 7.87 & 29.0\% & $+0.28$ \\
Spider test & Llama-3.3-70B & 80.02 & 80.76 & 80.53 & \textbf{81.04} & $+1.02$ & 87.70 & 7.69 & 13.3\% & $-0.23$ \\
\midrule
BIRD dev & GPT-4.1 & 65.71 & 67.21 & 66.62 & \textbf{69.43} & $+3.72$ & 78.03 & 12.32 & 30.2\% & $+1.30$ \\
BIRD dev & Qwen2.5-Coder-32B & 60.63 & 62.45 & 61.93 & \textbf{63.95} & $+3.32$ & 74.25 & 13.62 & 24.4\% & $+0.20$ \\
BIRD dev & Llama-3.3-70B & 57.50 & 58.87 & 58.28 & \textbf{59.84} & $+2.35$ & 71.25 & 13.75 & 17.1\% & $+0.20$ \\
\midrule
WikiSQL test & GPT-4.1 & 90.50 & 90.90 & 90.80 & \textbf{91.60} & $+1.10$ & 95.50 & 4.99 & 21.9\% & $+0.40$ \\
\bottomrule
\end{tabular}
\caption{EX (\%) of the four arms across backbones and benchmarks, each with
its own fixed three-candidate pool; derived columns computed from integer
counts. Headroom: raw selection misses over $n$ (pp); Recovered: share of
those misses BASIL-W removes; $\hat\gamma$: interaction point estimate (pp),
no interval.}
\label{tab:backbones}
\end{table*}

\paragraph{Main result.}
With GPT-4.1, BASIL-W reaches 88.50\% EX on Spider and 91.60\% on WikiSQL
(Table~\ref{tab:factorial}), gains of $+2.52$ and $+1.10$ percentage points
over raw evidence on
identical candidates and observations; each single factor helps less
(metadata $+1.12$ / $+0.40$; instruction $+0.75$ / $+0.30$), and all six
paired contrasts survive Holm correction (weakest $p=0.0054$).

\paragraph{Backbones and BIRD.}
Table~\ref{tab:backbones} and Figure~\ref{fig:results}a repeat the four arms
with two open-weight models and on BIRD dev. The ordering raw $<$
instruction $<$ metadata $<$ BASIL-W holds in all seven settings; these
comparisons are descriptive where paired tests are not reported. Gains are
largest on BIRD ($+3.72$, $+3.32$, $+2.35$), where raw headroom is 12--14
points, and smallest for Llama-3.3-70B (13--17\% of headroom recovered;
22--31\% elsewhere). Remaining errors split almost evenly: on Spider with
GPT-4.1, 129 of the 247 remaining errors have no correct candidate, while
118 have a correct candidate that the selector does not choose.
$\hat\gamma$ is positive in six settings and $-0.23$ for Llama-3.3-70B on
Spider, with no paired interval. The combined interface has the highest
observed accuracy, but these estimates do not establish synergy.

\paragraph{Selection baselines.}
Table~\ref{tab:baselines} compares four additional selectors on the same
candidate pools with GPT-4.1. Adding raw observations to the probe-free judge
gives observed gains of $+0.56$, $+0.06$ and $+0.65$ points. Execution-majority
and an MCS-SQL-style prompt score up to 2.0 points above the raw-evidence
judge. BASIL-W has the highest observed EX, with margins of $+1.58$ over
the strongest baseline on Spider and $+1.76$ on BIRD. These baseline
contrasts have no paired uncertainty estimates. The WikiSQL margin over
majority voting is $+0.15$; we claim neither superiority nor equivalence
there.

\paragraph{Absence-claim audit.}
Figure~\ref{fig:results}b and Table~\ref{tab:audit} report the Spider audit.
The raw and BASIL-W arms produced 600 and 510 audited assertions;
72 (12.00\%) and 22 (4.31\%) were labelled unsupported, affecting 59/400
(14.8\%) and 20/400 (5.0\%) questions. Inter-annotator agreement ranged from
$\kappa=0.81$ to $0.87$. Both the Spider and WikiSQL assertion labels and
aggregate audit statistics were verified under the predicate-matched
support criterion. The WikiSQL rates remain descriptive of the collected
assertions.

\paragraph{Comparison with published systems.}
Figure~\ref{fig:results}c and Table~\ref{tab:published} place BASIL-W
2.7 points below SIRIUS-SQL and 1.1--1.3 below the next four Spider systems,
which differ in candidate budget, training, and backbone. The pool's
oracle@3 of 93.99\% exceeds the highest listed score, so the candidate pool
does not by itself explain the performance gap. These reference scores
provide context and do not constitute matched comparisons.

\section{Conclusion}
BASIL-W makes the scope of database evidence explicit during fixed-pool SQL
selection. The combined interface achieved higher execution accuracy than
raw evidence or either factor alone in all seven evaluated settings, with
observed gains over four matched selection baselines on Spider and BIRD.
The verified audits showed lower unsupported absence-claim rates with
BASIL-W; the WikiSQL rates describe the collected assertions only.
On Spider with GPT-4.1, the remaining errors were almost evenly divided
between candidate generation and selection. The design does not isolate
scope understanding from prompt length or presentation, and the audit does
not establish a causal link between corrected assertions and improved
selections.

\section*{Limitations}
\paragraph{Evidence and statistical inference.}
The Spider and WikiSQL audit statistics were verified under the
predicate-matched support criterion. However, aggregate audit rates do not
show whether corrected assertions occurred on the questions whose
selections improved, and do not establish a causal mechanism.
Question-level linkage between assertion errors and selection gains was
not evaluated. The WikiSQL audit's original sampling procedure and number
of distinct questions were not recorded; its rates describe the collected
assertions and do not support population estimates. Differences in
preview-truncation frequency across benchmarks were not measured.

Paired significance tests cover the six GPT-4.1 factorial contrasts.
These tests treat questions as independent, although questions sharing a
database may be correlated. Cluster-aware confidence intervals are not
reported. The baseline comparisons, open-weight and BIRD settings, and
interaction estimates have no paired uncertainty intervals. Their observed
margins do not establish statistical superiority, equivalence, or synergy;
in particular, the $+0.15$ WikiSQL baseline margin remains descriptive.

\paragraph{Design and generalization.}
The factorial varies metadata and the explicit instruction separately,
but does not distinguish structured scope information from highlighting,
additional prompt tokens, or presentation. No minimal-representation
control was evaluated, so the results do not establish that every component
is necessary or minimal. Fixed candidate pools and cached observations
support matched comparisons but do not rule out benchmark exposure or
prompt-tuning bias. Public test questions may appear in pretraining.
BIRD results are on its development split, and comparisons with published
systems differ in backbone, training, and candidate budget.

\paragraph{Reproducibility and cost.}
The accompanying package provides the factorial prompt builder, witness
serialization, reporting scripts, and aggregate result tables. Complete run
manifests, full selector and baseline prompts, per-question model and scorer
records, and raw annotation records are not included, which limits
independent reproduction from this package. The package also omits the
complete prompt-tuning history and the Spider sampling script and audit
seed. The GPT-4.1 endpoint does not disclose an immutable model snapshot.
Scope serialization adds no probes, but adds input tokens and processing
and may affect output length. Latency and cost were not measured.

\paragraph{Scope of the contract.}
Scope metadata can be incorrect when traces, parser assumptions, or database
snapshots are inconsistent. A valid contract does not guarantee correct
selection, and the audit covers explicit assertions rather than unexpressed
reasoning. Original-database execution agreement can hide semantic query
errors. The SQLite pipeline does not establish transfer to Spider 2.0's
multi-engine workflows.

\bibliography{references}
\end{document}